\documentclass[conference]{IEEEtran}
\IEEEoverridecommandlockouts
\usepackage{cite}
\usepackage{amsmath,amssymb,amsfonts}
\usepackage{algorithmic}
\usepackage{graphicx}
\usepackage{textcomp}
\usepackage{xcolor}
\usepackage{url}
\usepackage{multirow}
\usepackage[vietnamese,english]{babel}
\usepackage[colorinlistoftodos]{todonotes}

\def\BibTeX{{\rm B\kern-.05em{\sc i\kern-.025em b}\kern-.08em
    T\kern-.1667em\lower.7ex\hbox{E}\kern-.125emX}}
\usepackage[numbers]{natbib}
\begin{document}

\title{Constructing a Knowledge Graph for Vietnamese Legal Cases with Heterogeneous Graphs
\\
}
\author{
Thi-Hai-Yen Vuong$^1$, Minh-Quan Hoang$^1$, Tan-Minh Nguyen$^1$, Hoang-Trung Nguyen$^1$, Ha-Thanh Nguyen$^2$ \\ 
\textit{$^1$VNU University of Engineering and Technology}, Hanoi, Vietnam \\ 
\textit{$^2$National Institute of Informatics}, Tokyo, Japan \\ 
$^1$\{yenvth, 21020553, 20020081, 20020083\}@vnu.edu.vn \\
$^2$nguyenhathanh@nii.ac.jp
}

\maketitle

\begin{abstract}
This paper presents a knowledge graph construction method for legal case documents and related laws, aiming to organize legal information efficiently and enhance various downstream tasks. Our approach consists of three main steps: data crawling, information extraction, and knowledge graph deployment. First, the data crawler collects a large corpus of legal case documents and related laws from various sources, providing a rich database for further processing. Next, the information extraction step employs natural language processing techniques to extract entities such as courts, cases, domains, and laws, as well as their relationships from the unstructured text. Finally, the knowledge graph is deployed, connecting these entities based on their extracted relationships, creating a heterogeneous graph that effectively represents legal information and caters to users such as lawyers, judges, and scholars. The established baseline model leverages unsupervised learning methods, and by incorporating the knowledge graph, it demonstrates the ability to identify relevant laws for a given legal case. This approach opens up opportunities for various applications in the legal domain, such as legal case analysis, legal recommendation, and decision support.
\end{abstract}


\section{Introduction}

Along with the development of technology, the volume of digital documents has significantly increased, especially in the legal field. We can easily search for and access legal information more efficiently. Legal documents are often lengthy, structured, and presented in a specific writing style. To effectively harness this data, much depends on how it is organized and standardized. In the legal domain, particularly in legal case documents, one can find information about the cases, court decisions, and laws related to those cases. Although the information is available, retrieving legal information can be complex, especially when dealing with specific legal cases or investigating a particular legal case as a legal expert. Desired information may need to be searched from various sources and approached in different ways.

Typically, searching for information within legal information systems is relatively straightforward, relying on keyword matching or querying metadata about the documents (such as document codes, types, court decisions, etc.). However, searching for legal cases or related laws often occurs manually, consuming time and effort. Due to the substantial number of legal cases and related laws, which are often lengthy, intricately structured, and contain numerous domain-specific terms, the search process may involve erroneous keyword selection, leading to undesired search outcomes.

Contextual search is implemented to address the limitations of keyword-based searches \cite{maxwell2008concept}. Nonetheless, lengthy queries pose a challenge in information retrieval. Both representation learning methods and matching learning methods have limitations in handling lengthy documents. Moreover, during the composition of documents within the court environment, legal judgments may contain typographical errors, punctuation inaccuracies, or may suffer from conversion issues when transitioning from PDF files. Additionally, the style of these legal cases is contingent upon the court clerk responsible for the record proceedings.

The method of constructing a knowledge graph serves as a suitable tool for identifying and representing the relationships between legal cases and relevant laws \cite{casanovas2016semantic,fernandez2011legal}. Knowledge graphs can effectively depict vast amounts of knowledge with semantic meaning, facilitating easy access and structured querying. These knowledge graphs are designed in a user-friendly manner, catering to non-expert users such as lawyers, judges, scholars, etc., enabling them to easily utilize and explore the information. Moreover, knowledge graphs can be applied to enhance various downstream tasks in the legal domain such as information retrieval \cite{crotti2020knowledge}, question-answering \cite{cui2019kbqa, sovrano2020legal}, classification \cite{ashley2017artificial}, and more.

In recent years, deep learning methods such as Convolutional Neural Networks (CNNs) \cite{Gao2014ModelingIW, hu2014convolutional}, Recurrent Neural Networks (RNNs) \cite{palangi2016deep}, language models \cite{devlin2018bert, beltagy2020longformer}, and large language models \cite{ni2021sentence,touvron2023llama} have demonstrated promising results in the field of natural language processing (NLP) in general, and specifically in the domain of legal natural language processing. In broader domains, labeled data is often abundant, whereas in the narrower legal domain, labeled datasets are relatively scarce. One of the reasons is legal texts are often subject to copyright restrictions and privacy concerns, which restrict the availability and sharing of annotated or labeled legal datasets. Additionally, legal texts require expert knowledge for accurate annotation and labeling, as legal concepts and nuances may be challenging for non-experts to interpret correctly. Annotating legal datasets requires legal expertise, which may limit the availability of annotators and increase the cost and time required for dataset creation. There are some public legal datasets in English \cite{rabelo2020summary}, Chinese \cite{huang2020named}, German \cite{leitner2019fine}, or Japanese \cite{rabelo2020summary}. However, using them can be challenging due to the variations in legal systems across different countries. In the case of Vietnam, the process of digitizing the legal system is underway, and creating a standardized dataset remains a significant challenge. Furthermore, generating a training dataset for supervised learning tasks could be time-consuming and costly.

This study aims to construct a knowledge graph for legal case documents and related laws and apply them to downstream tasks. The main contributions of this paper are as follows:
\begin{itemize}
    \item Building a dataset of legal case documents and relevant laws.
    \item Analyzing, processing, and extracting information to construct the knowledge graph.
    \item Establishing a baseline model using unsupervised learning methods and knowledge graph for identifying relevant laws.
\end{itemize}

The paper is structured as follows: In Section 2, we present related works. A description of the knowledge graph is provided in Section 3. Section 4 outlines the steps to construct the knowledge graph. Section 5 presents the baseline model for identifying relevant laws, and finally, we conclude the paper.

\section{Related Works}
In the early stages of natural language processing research in the legal domain, foundational efforts were dedicated to rule-based systems \cite{susskind1986expert,rissland1988artificial}. These systems, encompassing expert systems or lexical matching approaches, provided valuable advancements by facilitating information retrieval and comprehension within the legal context. In recent times, machine learning methods \cite{bench2012history,mandal2017measuring}, particularly deep learning \cite{tran2020encoded,nguyen2022attentive}, have also been applied to address challenges in automated legal text processing.

The establishment of knowledge graphs has gained substantial attention as an effective approach to represent and organize legal information. \citeauthor{filtz2017building} explores data representation and search within the legal domain, proposing an approach to construct a legal knowledge graph using legal data from Austria \cite{filtz2017building}. \citeauthor{tang2020salkg} introduces SALKG, a semantic annotation system designed to construct a high-quality legal knowledge graph through a semi-automatic approach \cite{tang2020salkg}. \citeauthor{sovrano2020legal} constructed an integrated Knowledge Graph based on combining open Knowledge extraction and natural language processing techniques, along with key ontology design patterns specific to the legal domain \cite{sovrano2020legal}. These patterns include event, time, role, agent, right, obligations, and jurisdiction. A question answering model has been developed from the legal knowledge graph to facilitate information retrieval and respond to these queries efficiently. \citeauthor{dhani2021similar} construct a knowledge graph to enhance the efficiency of finding similar legal cases \cite{dhani2021similar}. 

\section{Problem Statement}
\begin{figure}[ht]
    \centering
    \includegraphics[width=0.45\textwidth]{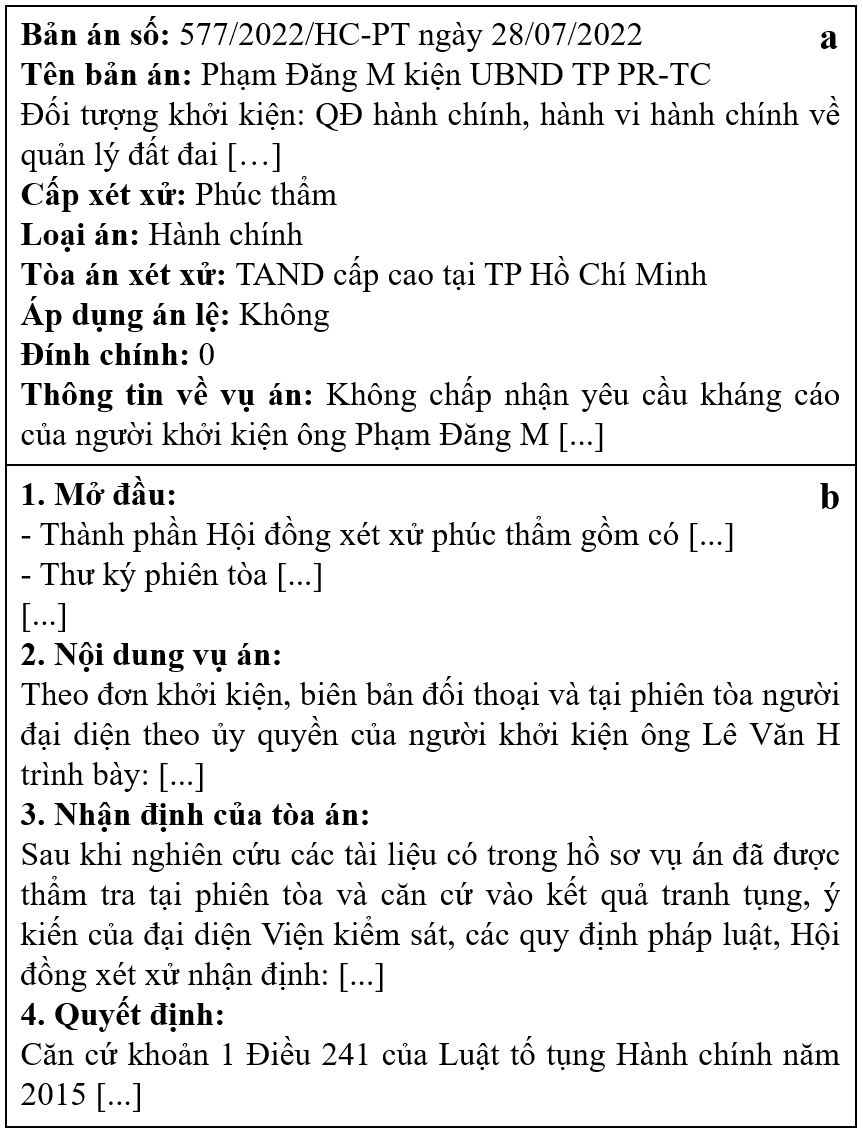}
    \caption{Structure of a legal case (a: meta-data, b: case document)}
    \label{fig:case_structure}
\end{figure}

A legal case archived on the website of the Vietnam Supreme People's Court consists of two parts: meta-data and case document. Figure \ref{fig:case_structure} illustrates the structure and content of a legal case. The meta-data contains basic information about a case including the case's number, case's name, type of case, etc. The body of the legal case document comprises four sections: the Introduction, the Content of the case, the Court's judgment, and the Court's decision. The description of each part is shown in Table \ref{tab:case structure}.

\begin{table}[ht]
\caption{Description of a law document.}
\label{tab:case structure}
\begin{tabular}{|p{2.5cm}|p{5.5cm}|}
\hline
\multicolumn{1}{|c|}{\textbf{Part}} & \multicolumn{1}{c|}{\textbf{Description}}                                                         \\ \hline
Introduction                  & details of case, court, defendant, plaintiff, related parties (e.g full name, date of birth, address of parties)      \\ \hline
Content of the case           & opinions of case, court, defendant, plaintiff, related parties      \\ \hline
Court's judgment & Opinions, analysis of the court \\ \hline
Court's decision & Decisions of the court based on above parts\\ \hline
\end{tabular}
\end{table}

In this work, we construct the Vietnamese legal knowledge graph based on a heterogeneous graph, which is a special kind of network for presenting multiple kinds of entities or multiple kinds of relations. A heterogeneous graph ${G = (V, E)}$ contains an entity set $V$ and a relation set $E$. The graph is also associated with a node type mapping function $f: V \rightarrow A$ and a relation type mapping function $g: E \rightarrow R$. $A$ and $R$ denote the sets of entity types and relations types, where $|A| + |R| > 2$.
Particularly, we define 4 types of entity based on the characteristic of the Vietnamese case law, including:
\begin{itemize}
    \item Case node, which embeds information about each judgment/trial that is currently in effect.
    \item Domain node, which embeds information about crimes, types of disputes and decisions.
    \item Court node embeds information about every court's name and level in the juridical system.
    \item Law node contains the name of specific law/code of law
\end{itemize}
There are a total of 3 types of relations between entities, including:
\begin{itemize}
    \item Decide relation between courts and cases, indicating the relationship of a particular court hearing the trial.
    \item Belong-to relation between cases and domains, indicating the relationship of a particular domain and subdomain under which the case falls.
    \item Based-on relation between cases and laws, indicating the relationship of a particular judgment or decision that has referenced a set of laws/codes of law to support its verdict.
\end{itemize}

In a heterogeneous graph, two entities can be connected through different semantic paths, which are called meta-paths.
A meta-path $P$ is defined as a path in the form of $A_1 \xrightarrow{R_1} A_2 \xrightarrow{R_2} \dots \xrightarrow{R_k} A_{k+1}$, which presents a composite relation $R=R_1 \circ R_2  \circ \dots \circ R_k$ between $A_1$ and $A_{k+1}$, where $\circ$ denotes the composistion operator on relations. Two legal cases can be connected via different meta-paths, e.g. Case-Court-Case (CCC) or Case-Domain-Case (CDC). Each meta-path reveals different semantics. For example, the CCC path means these cases were judged by the same court, while the CDC path denotes that they belong to the same domain.


\section{Building Knowledge Graph}
\begin{figure*}
    \centering
    \includegraphics[width=0.48\paperwidth]{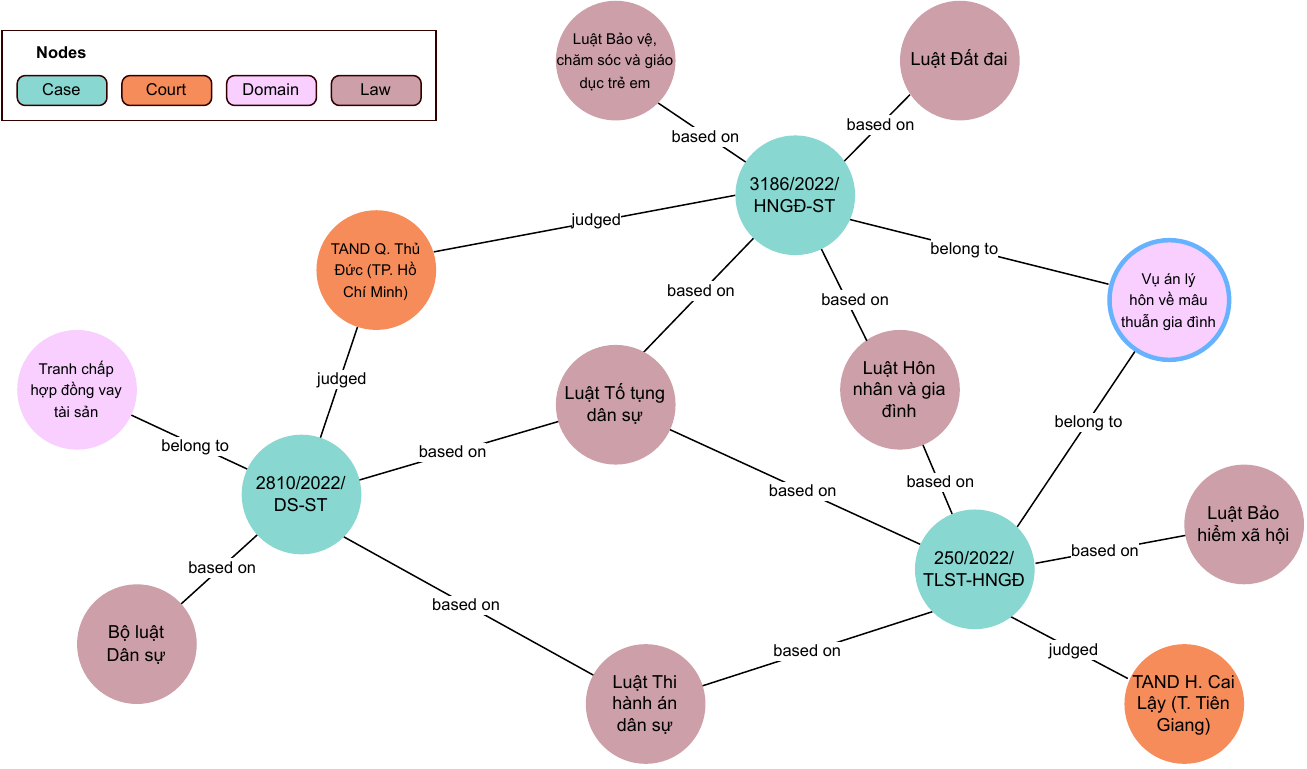}
    \caption{The knowledge graph visualization}
    \label{fig:visualize_graph}
\end{figure*}

In this section, we describe our approach for constructing a knowledge graph for Vietnamese legal cases. Each of the following subsections present steps of our approach in detail. Figure \ref{fig:visualize_graph} visualize a portion of the graph with a case node in cyan, a court node in orange, a domain node in pink, and a law node in brown. 

\subsection{Data Crawler}
\selectlanguage{vietnamese}
The database contains 9578 court cases published by the Supreme People's Court of Vietnam in plain text and 225 valid law codes of Vietnam. 
 
The name of 225 valid laws/codes of Vietnam has been crawled from the website\footnote{\url{https://thuvienphapluat.vn}}. The court cases have been scraped using a Python-based engine from the website\footnote{\url{https://congbobanan.toaan.gov.vn}} that archives many legal documents from various courts and tribunal repositories. The database comprises court cases for cases from many legal domains such as criminal law, civil law,  marriage and family law, etc.). Crawled cases contains two parts including meta-data and case document. 

\subsection{Information Extraction}
Information extraction is performed to extract information on entities and relations from a legal case document. All the entities are extracted to form useful information about the context of the case. Table \ref{tab:entity} presents the list of entities in the legal case document. 

Entities such as court, domain, and case are extracted from both meta-data and case document parts using regular expressions. However, the meta-data path lacks information on the laws/codes involved in the case. To retrieve these laws/codes, we first extracted sentences containing information of laws/codes using regular expression. These sentences may contain noise and redundant information. To address this, we proceed to match these sentences with our database of 225 Vietnam laws/codes. An illustration of the Law entity extraction step is presented in Table \ref{tab:law_ext}. 

\selectlanguage{english}
\begin{table}[ht]
\caption{Entities and their atributes to annotate data}
\label{tab:entity}
\begin{tabular}{|p{1cm}|p{2cm}|p{4.5cm}|}
\hline
\multicolumn{1}{|c|}{\textbf{Entity}} & \multicolumn{1}{c|}{\textbf{Attributes}} & \multicolumn{1}{c|}{\textbf{Description}}                      \\ \hline
\multirow{9}{*}{Case}                 & case\_id                                 & id of the case                                                 \\ \cline{2-3} 
                                      & case\_number                             & number of the case (e.g 577/2022/HC-PT)                         \\ \cline{2-3} 
                                      & document\_type                           & type of the document (Verdict or Decision)                     \\ \cline{2-3} 
                                      & case\_level                              & level of the court (Trial, Appellate, and Cassation/Reopening) \\ \cline{2-3} 
                                      & case\_content                            & basic information of the case                                  \\ \cline{2-3} 
                                      & case\_text                               & full content of the case                                       \\ \cline{2-3} 
                                      & date                                     & documented and relevant dates                                  \\ \cline{2-3} 
                                      & court\_id                                & id of the court                                                \\ \cline{2-3} 
                                      & domain\_id                               & id of the case's domain                                    \\ \hline
\multirow{3}{*}{Domain}               & domain\_id                               & id of the domain                                               \\ \cline{2-3} 
                                      & domain\_name                             & type of the case (e.g Criminal, Civil, etc)                \\ \cline{2-3} 
                                      & subdomain                                & crimes, legal relations in the domain                          \\ \hline
\multirow{3}{*}{Court}                & court\_id                                & id of the court                                                \\ \cline{2-3} 
                                      & court\_name                              & name of the court (e.g Hanoi Supreme People's Court)           \\ \cline{2-3} 
                                      & court\_level                             & level of the court (e.g Provincial People's Court)             \\ \hline
\multirow{2}{*}{Law}                  & law\_id                                  & id of the law                                                  \\ \cline{2-3} 
                                      & law\_name                                & name of the law (e.g Criminal Code, Civil Code)                \\ \hline
\end{tabular}
\end{table}

\begin{table}[ht]
\caption{Law entity extraction in a legal case document.}
\selectlanguage{vietnamese}
\label{tab:law_ext}
\begin{tabular}{|p{4cm}|p{4cm}|}
\hline
\multicolumn{1}{|c|}{\textbf{Extracted sentence}} & \multicolumn{1}{c|}{\textbf{Laws/Codes corpus}} \\ \hline
Điều 19 của Luật Hôn nhân và Gia đình \textit{(Article 19 of the Marriage and Family Law) }                                  & Luật Thi hành án dân sự sửa đổi 2014 \textit{(Amended Civil Judgment Enforcement Law 2014)}     \\ \hline
điều 81, 82 và 83 của Luật Hôn nhân và Gia đình \textit{(Articles 81, 82, and 83 of the Marriage and Family Law)}                          & Luật thi hành án dân sự 2008 (Civil Judgment Enforcement Law 2008)             \\ \hline
khoản 1 Điều 51, các điều 56, 81, 82 và 83 của Luật Hôn nhân và Gia đình \textit{(Clause 1 of Article 51, Articles 56, 81, 82, and 83 of the Marriage and Family Law)} & Luật tổ chức Tòa án nhân dân 2014 \textit{(People's Court Organization Law 2014)}        \\ \hline
điều 28, 35, 39, 147, 227, 228 và 273 của Bộ luật Tố tụng dân sự \textit{(Articles 28, 35, 39, 147, 227, 228, and 273 of the Civil Procedure Code) }        & Luật thi hành án hình sự 2010 (\textit{Criminal Judgment Enforcement Law 2010)}           \\ \hline
điều 6, 7, 7a và 9 của Luật Thi hành án dân sự \textit{(Articles 6, 7, 7a, and 9 of the Civil Judgment Enforcement Law)}                          & Bộ luật Tố tụng dân sự 2004  \textit{(Civil Procedure Code 2004)   }         \\ \hline
Điều 30 của Luật Thi hành án dân sự \textit{(Article 30 of the Civil Judgment Enforcement Law)}                                      & Luật Hôn nhân và gia đình 2014 \textit{(Marriage and Family Law 2014)}           \\ \hline
\end{tabular}
\end{table}
 


\subsection{Knowledge Graph Deployment}
Table \ref{tab:stat} shows the statistics of the legal knowledge graph. It has a total of 10181 nodes, out of which 9078 are case nodes. The total number of edges is 54110 edges, 35954 of that are between case nodes and law nodes, while both the relations of (case, court) and (case, domain) pairs have 9078 edges. This is due to the one-to-one link among case nodes, domain nodes and court nodes. The density $D$ of the graph is 0.001, and the ratio $R$ of the number of edges per node is 5.314. These scores are calculated by the Formula \ref{eq:density} and Formula \ref{eq:ratio}, respectively. 

\begin{table}[ht]
\selectlanguage{english}
\caption{Statistics of the graph}
\label{tab:stat}
\centering
\begin{tabular}{|l|c|}
\hline
\multicolumn{1}{|c|}{\textbf{Property}} & \multicolumn{1}{c|}{\textbf{Quantity}} \\ \hline
Case node                           & 9078                                   \\
Court node                              & 693                                    \\
Domain node                             & 185                                    \\
Law node                                & 225                                    \\
Total                                   & 10181                                  \\ \hline
Case-law edge                       & 35954                                  \\
Case-domain edge                    & 9078                                   \\
Judgement-court edge                    & 9078                                   \\
Total                                   & 54110                                  \\ \hline
Connected components                    & 60                                    \\ \hline
\end{tabular}
\end{table}

\begin{equation}
    D = \frac{|E|}{|V|\times(|V| - 1)}
    \label{eq:density}
\end{equation}
\begin{equation}
    R = \frac{|E|}{|V|}
    \label{eq:ratio}
\end{equation}
where $|E|$ is the number of edge and $|V|$ is the number of vertex in the graph.

One interesting insight of the graph is connectivity. Although the graph has 60 connected components, only one of them is significant, all others only have 1 node. This is the result of some law nodes do not link to any of the case nodes in the crawled data.
The significant component has 10122 nodes, with 9078 nodes representing cases and 176 nodes representing laws. Further analysis reveals that there are approximately 4 based-on relations per case node, which means for each case, 4 laws/codes of law are referenced.
This connected component indicates that there is a strong relation among all case nodes via different meta-paths.

\section{Relevant law determiner baseline and results}
\subsection{Relevant law determiner baseline}

BM25 Okapi is a powerful lexical engine used for ranking a collection of documents based on the frequency of query terms in each document. 
Given a query $Q$, containing keywords $q_1,q_2,\dots,q_n$, the BM25 score of a document $D$ is computed by Formula \ref{eq:bm25}.
\begin{equation}
    \footnotesize
    score(D,Q)=\sum_{i=1}^n IDF(q_i) \times \frac{f(q_i,D)\times(k_1+1)}{f(q_i,D)+k_1\times (1-b+b\times \frac{|D|}{avgdl})}
    \label{eq:bm25}
\end{equation}
where $f(q_i,D)$ is the number of times that $q_i$ occurs in the document $D$, $|D|$ is the length of document $D$ in words, and $avgdl$ is the average document length in the text collection, $k1$ and $b$ are free parameters.

There are two approaches to aggregating relevant law sets of different cases: union and intersection. Given two law sets $A$ and $B$, the union set an intersection set of $A$ and $B$ are defined as follow:
\[A \cup B = \{x : x \in A \text{ and } x \in B\}\]
\[A \cap B = \{x : x \in A \text{ or } x \in B\}\]
If $A \cap B = \emptyset$ then the system would return an empty list.

We implement four methods based on the BM25 search engine for the task of determining relevant laws as follows:
\begin{enumerate}
    \item The first method is called \textit{case-law matching}, which means each part of a case is directly fed into the BM25 engine to retrieve top-1 relevant law.
    \item The second method is \textit{improved case-law matching}. We combine the results from the first method (run 1, 2, and 3 in Table \ref{tab:result}) using union and intersection aggregate functions.
    \item The third method is \textit{case-case matching and KG}. We first search for top-2 similar cases in the corpus using BM25 engine. The relevant laws of candidate cases are extracted via the knowledge graph. Finally, these laws are aggregated using the union and intersection function to produce relevant laws of the query case.
    \item The last method is \textit{Domain case-case matching and KG}. Instead of querying similar cases in the whole dataset as in method 3, search space is reduced via the meta-path Case-Domain-Case in the knowledge graph. Particularly, domain-related cases are extracted to form a candidate list. We then extract top-2 similar cases from this list using the BM25 engine.
    The relevant laws extraction step is similar to method 3.
\end{enumerate}

\subsection{Experiments and Results}
In this work, a baseline model based on the BM25 engine is applied to evaluate the KG application in the law retrieval task. 
Particularly, we performed 11 runs of 4 methods on the test set of 500 legal cases. The heterogeneous graph does not contain these test cases. Table \ref{tab:result} presents the details of experimental results. 

The runs of methods 1 and 2 have limited results. One of the reasons is the length of both the query case and the law and furthermore, the vocabulary correlation between them is not substantial.
For the third run, the content of the court's decision part is input into the BM25 engine to retrieve relevant laws/codes. This part refers to a lot of laws/codes information to support decisions. As a result, the third run achieves the highest precision of 0.676.

The methods using knowledge graphs achieve outstanding results. In the tenth run, similar cases are extracted via the meta-path Case-Domain-Case in the knowledge graph, which achieves the highest F1 of 0.503 and Recall of 0.583. Compared to methods 1 and 2, the utilization of legal knowledge graphs achieves a remarkable increase of 21\% in F1 score. Subsequently, the combination of domain-specific information from the legal knowledge graph also contributes to reducing search space and increasing accuracy, thereby improving the model's performance.

\begin{table}[ht]
\caption{Relevant laws retrieval results}
\label{tab:result}
\centering
\begin{tabular}{|c|l|c|c|c|}
\hline
\multicolumn{1}{|c|}{\textbf{\#}} & \multicolumn{1}{c|}{\textbf{Description}} & \multicolumn{1}{c|}{\textbf{F1}}    & \multicolumn{1}{c|}{\textbf{Recall}} & \textbf{Precision} \\ \hline
\multicolumn{5}{|l|}{\textit{Method 1 - Case-law matching}}                                                                                                                                                       \\ \hline
\multicolumn{1}{|c|}{1}           & \multicolumn{1}{l|}{Content of the case}                & \multicolumn{1}{c|}{0.061}          & \multicolumn{1}{c|}{0.037}           & 0.18               \\ \hline
\multicolumn{1}{|c|}{2}           & \multicolumn{1}{l|}{Court's judgment}                    & \multicolumn{1}{c|}{0.154}          & \multicolumn{1}{c|}{0.093}           & 0.093              \\ \hline
\multicolumn{1}{|c|}{3}           & \multicolumn{1}{l|}{Court's decision}                   & \multicolumn{1}{c|}{0.231}          & \multicolumn{1}{c|}{0.139}           & \textbf{0.676}     \\ \hline
\multicolumn{5}{|l|}{\textit{Method 2 - Improved case-law matching}}                                                                                                                                                       \\ \hline
\multicolumn{1}{|c|}{4}           & \multicolumn{1}{l|}{Mix 3 queries (Union)}              & \multicolumn{1}{c|}{0.288}          & \multicolumn{1}{c|}{0.347}           & 0.245              \\ \hline
\multicolumn{1}{|c|}{5}           & \multicolumn{1}{l|}{Mix 3 queries (Intersection)}       & \multicolumn{1}{c|}{0.029}          & \multicolumn{1}{c|}{0.015}           & 0.321              \\ \hline
\multicolumn{5}{|l|}{\textit{Method 3 - Case-case matching and KG}}                                                                                                                                                       \\ \hline
\multicolumn{1}{|c|}{6}           & \multicolumn{1}{l|}{Top-1 similar case}                 & \multicolumn{1}{c|}{0.449}          & \multicolumn{1}{c|}{0.429}           & 0.472              \\ \hline
\multicolumn{1}{|c|}{7}           & \multicolumn{1}{l|}{Top-2 similar cases (Union)}        & \multicolumn{1}{c|}{0.471}          & \multicolumn{1}{c|}{0.554}           & 0.409              \\ \hline
\multicolumn{1}{|c|}{8}           & \multicolumn{1}{l|}{Top-2 similar cases (Intersection)} & \multicolumn{1}{c|}{0.386}          & \multicolumn{1}{c|}{0.281}           & 0.616              \\ \hline
\multicolumn{5}{|l|}{\textit{Method 4 - Domain case-case matching and KG}}                                                                                                                                                       \\ \hline
\multicolumn{1}{|c|}{9}           & \multicolumn{1}{l|}{Top-1 similar case}                 & \multicolumn{1}{c|}{0.47}           & \multicolumn{1}{c|}{0.442}           & 0.503              \\ \hline
\multicolumn{1}{|c|}{10}          & \multicolumn{1}{l|}{Top-2 similar cases (Union)}        & \multicolumn{1}{c|}{\textbf{0.503}} & \multicolumn{1}{c|}{\textbf{0.583}}  & 0.441              \\ \hline
\multicolumn{1}{|c|}{11}          & \multicolumn{1}{l|}{Top-2 similar cases (Intersection)} & \multicolumn{1}{c|}{0.411}          & \multicolumn{1}{c|}{0.303}           & 0.642              \\ \hline
\end{tabular}
\end{table}
Furthermore, experiments show that there is a trade-off between F1, Recall and Precision scores when using Union and Intersection aggregation approaches. 
Run 10 returns related laws that supports at least one similar case. As a result, this run has a higher recall. Meanwhile, Run 11 only retains laws that supports all the similar cases, which leads to a higher precision. 

For error analysis, Table \ref{tab:output examples} presents output examples from Run 10 and Run 11 (matching using KG). Run 10 returns 7 laws/codes, in which the first 2 laws/codes are correct. Although the result contains all related laws in the given legal case, its precision score is low due to an excessive number of retrieved cases.    
Compared to Run 10, Run 11 only returns 3 laws/codes, in which one of them is correct. Therefore, this run achieves higher precision and lower recall scores.
\begin{table}[ht]
\caption{Output examples of Run 10 and Run 11}
\selectlanguage{vietnamese}
\label{tab:output examples}
\begin{tabular}{|p{4cm}|p{4cm}|}
\hline
\multicolumn{1}{|c|}{\textbf{Run 10}}                                              & \multicolumn{1}{c|}{\textbf{Run 11}}                                     \\ \hline
\textbf{Luật Hôn nhân và gia đình 2014 (Marriage and Family Law 2014)}           & \textbf{Luật Hôn nhân và gia đình 2014 (Marriage and Family Law 2014)} \\ \hline
\textbf{Bộ luật tố tụng dân sự 2015 (Civil Procedure Code 2015) }                           & Luật thi hành án dân sự 2008 (Civil Judgment Enforcement Law 2008)       \\ \hline
Luật phí và lệ phí 2015 (Fees and Charges Law 2015)                                & Bộ luật tố tụng hình sự 2015 (Criminal Procedure Code 2015)              \\ \hline
Luật Thi hành án dân sự sửa đổi 2014 (Amended Civil Judgment Enforcement Law 2014) &                                                                          \\ \hline
Luật Bảo hiểm xã hội 2014 (Law on Social Insurance 2014)                           &                                                                          \\ \hline
Luật thi hành án dân sự 2008 (Civil Judgment Enforcement Law 2008)                 &                                                                          \\ \hline
Bộ luật tố tụng hình sự 2015 (Criminal Procedure Code 2015)                        &                                                                          \\ \hline
\end{tabular}
\end{table}
\selectlanguage{english}

\section{Conclusions}
This paper proposes a novel approach to construct a knowledge graph for legal case documents and related laws, enhancing the organization and representation of legal information. Our method involves data crawling, information extraction, and knowledge graph deployment, effectively extracting entities and relationships from unstructured text and forming a heterogeneous graph. This approach enables various applications in the legal domain, such as legal case analysis, legal recommendation, and decision support. The established baseline model, which leverages unsupervised learning methods and the knowledge graph, demonstrates promising results in identifying relevant laws for a given legal case. Future work can focus on improving information extraction, incorporating advanced graph-based learning techniques, and expanding the scope of the knowledge graph for better performance and broader applicability.



\bibliographystyle{IEEEtranN}
\bibliography{ref}

\end{document}